\documentclass[letterpaper]{article} % DO NOT CHANGE THIS
\usepackage{aaai2027}  % DO NOT CHANGE THIS
\usepackage[hyphens]{url}  % DO NOT CHANGE THIS
\usepackage{graphicx} % DO NOT CHANGE THIS
\usepackage{natbib}  % DO NOT CHANGE THIS
\usepackage{caption} % DO NOT CHANGE THIS
\usepackage{enumitem}
\usepackage{booktabs}   % better tables
\usepackage{amsmath}    % metric definitions
\usepackage{xcolor}     % ONLY for the outline notes below; remove later

\title{Clinician-Grounded Quality Assurance for AI-Assisted Psychiatric Intake}
\author{
    King Shi\textsuperscript{\rm 1},
    Amanda Li\textsuperscript{\rm 1},
    Jonathan Ivey\textsuperscript{\rm 1},
    Synthia Qia Wang\textsuperscript{\rm 1},
    Guan Gui\textsuperscript{\rm 1},
    Hyunseo Kim\textsuperscript{\rm 1},
    Peter Zandi\textsuperscript{\rm 1},
    Jason Straub\textsuperscript{\rm 1},
    Jacob Taylor\textsuperscript{\rm 1},
    Ananya Joshi\textsuperscript{\rm 1}
}
\affiliations{
    \textsuperscript{\rm 1} Johns Hopkins University\\
}

\begin{document}
\maketitle

%% ============================================================
\begin{abstract}
Before patients can use AI-assisted psychiatric intake systems, health systems need practical ways to routinely evaluate these tools against their clinical standards for quality assurance. Because clinicians may use different intake styles, evaluation for this task must (1) support comparison across interviewing approaches, (2) minimize clinician burden, and (3) measure clinically relevant performance for health systems deploying these technologies. We present a clinician-grounded evaluation platform built around a memory-augmented patient simulator for open-ended AI interviewing, InterviewPlayground. We created interactive patients using InterviewPlayground with our expert-authored vignettes, constructed a simulated intake platform for the interviews, and designed evaluation modalities relevant to intake. In a pilot of 6 clinicians in a 25-minute assessment compared to a GPT-based LLM intake interviewer, the LLM recovered more of the clinically relevant items embedded in the patient vignettes (88.0\% vs. 38.9\%), but made more clinical inferences not based on the interview (56.8\% vs. 27.8\%), and characterized identified safety concerns less often (33.3\% vs. 66.7\%), setting the stage for deployed quality assurance for this task. 
\end{abstract}

%% hero figure
\begin{figure*}[t]
\centering
\includegraphics[width=\textwidth]{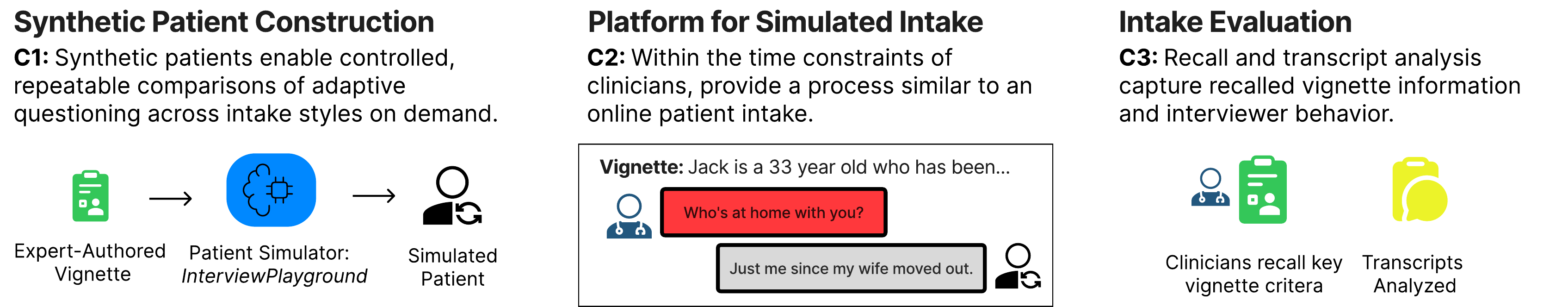}
\caption{Overview of the clinician-grounded quality-assurance platform for AI-assisted psychiatric intake. The platform combines repeatable simulated patients for cross-interviewer comparison (C1), a low-burden clinician intake workflow (C2), and a form and transcript-based evaluation of clinically relevant performance (C3).}
\label{fig:pipeline}
\end{figure*}

\section{Introduction and Motivation}

Psychiatric intake is the first structured clinical conversation between a clinician and a patient. During intake, a clinician identifies the chief complaint, assesses the risk of self-harm, characterizes psychiatric and medical symptoms and their severity, and determines the next steps~\cite{silverman2015}. This is demanding work under tight time constraints, and at Johns Hopkins, intakes are routinely longer than an hour. Not everyone has access to this type of care -- nearly 33\% of the U.S. population lives in a mental health professional shortage area \cite{hrsa2026hpsa}, and missed information during intake can contribute to delayed diagnosis and treatment, particularly for frequently misdiagnosed conditions \cite{hirschfeld2003}. 

AI-assisted psychiatric intake has the potential to significantly improve the standard of psychiatric care by reducing the amount of time needed for information gathering and allowing a clinician to focus on building rapport with the patient, developing an accurate diagnosis, and building a treatment plan. Recent advances in automated interviewing and AI-based mental health support have prompted professional organizations and startups to explore chatbots for psychiatric intake (e.g. Form Doctor\footnote{https://formdr.com} and Osmind\footnote{https://www.osmind.org}).

%% However, psychiatric assessment is difficult to scale. An average psychiatric visit lasts about 38 minutes \cite{olfson1999}, while a complex intake may take several hours. Asynchronous AI-assisted intake could expand access, particularly f

Before patient-facing deployment, health systems need credible ways to determine whether AI-assisted intake systems meet local clinical standards. Directly testing an unvalidated system with psychiatric patients creates numerous ethical and clinical risks \cite{apa2025advisory}, and clinical-AI reporting standards call for structured evaluation before patient contact \cite{vasey2022,gallifant2025}. This is important for deployment because health systems need quality-assurance processes that can verify AI intake tools operate reliably within clinical workflows before patient-facing use and as models, prompts, and clinical standards change.

Still, psychiatric intake is difficult to assess systematically because it is an adaptive conversation. Clinicians vary in the way they conduct an intake by style, and the appropriate next question depends on the patient's preceding response \cite{Oneilll2021,silverman2015}. Quality assurance must account for different interviewing styles while checking that important information is gathered and patient disclosures are handled appropriately. In developing a quality-assurance approach for AI-assisted psychiatric intake, our interdisciplinary precision-psychiatry team at Johns Hopkins identified three challenges.%%Nevertheless, effective intakes often recover the same clinically important information, and even if clinicians take different paths to get there, they should always respond appropriately to the patient. %%Evaluation must therefore measure more than whether an AI system mimics a preferred interviewing style or covers a checklist of topics.

\textbf{C1: Meaningful Comparisons Across Styles} Quality assurance criteria cannot be overly sensitive to differences in style,  and also needs to be comparable and replicable between clinicians. Historical transcripts are difficult to use to train a standardized evaluator because the same patient is rarely interviewed by multiple clinicians, and intake notes vary substantially in style. In contrast, standardized patient actors can present the same case to multiple clinicians, but can be resource intensive to gather several clinicians simultaneously and performance can vary between encounters \cite{flanagan2023}, and even subtle changes in presentation change the interview. We address this by creating synthetic patients with fixed clinical ground truth fields in vignettes that can be interviewed on-demand by different clinicians or AI systems.

\textbf{C2: Low Clinician Burden} Clinicians do not have the bandwidth to regularly stress-test and tune AI intake systems for quality assurance. We needed a platform for clinicians to efficiently provide reference interviews reflecting local intake practice. These sessions needed to be sufficiently long to allow clinicians to conduct the intake process consistent with their experience, while remaining sufficiently brief to facilitate the completion of the session. The simulated platform must also be realistic and accessible for clinicians to provide accurate data points. We therefore implement a recall form to capture what the interviewer retained from the encounter, alongside transcript analysis of information elicitation, interviewer behavior, inference, and responses to safety concerns.

\textbf{C3: Deployment-Relevant Evaluation} Determining whether an intake was adequate requires knowing both what was elicited during the conversation and what the interviewer took away from it, which goes beyond just the transcript generated from the intake process. We therefore implement a recall form, which focuses on whether certain clinical information was elicited during the interview by the clinician, and a transcript analysis to evaluate interviewer behavior, inference, and responses to safety concerns.

Our approach, shown in Figure~\ref{fig:pipeline}, is designed to address these challenges, and we conducted a pilot study with practicing clinicians (n=6) and a candidate AI psychiatric intake application. First, synthetic patients generally remained consistent across repeated simulated intake interviews. Second, clinicians completed the workflow in a median 25.3 minutes and 6 completed the full task despite the substantial time demands of clinician participation. Third, the evaluation extended beyond binary clinical information recall, and revealed nuanced limitations needed for quality assurance: while the LLM recovered more positive ground-truth fields (88\% vs. 39\%), it made more unsupported inferences (57\% vs. 28\%) and characterized identified safety concerns less often (33\% vs. 67\%). This pilot establishes an approach for health systems that need a repeatable, clinician-grounded process for quality assurance around AI-assisted psychiatric intake.

\section{Related Work}

Today, conversational AI systems for behavioral health are being developed faster than quality-assurance standards and evaluation guidelines can keep pace \cite{RAY2023, hua2025}, posing considerable challenges for clinical deployment. Research has shown that LLMs can express stigma toward psychiatric conditions and respond inappropriately to disclosures of suicidal ideation \cite{moore2025, pichowicz2025}, and health systems have concerns about these technologies being patient-facing. Existing psychiatric LLM benchmarks are largely static and focus on model knowledge, whereas intake requires evaluating how a model conducts an adaptive interview \cite{fouda2026,presacan2026,liu2025}.

\paragraph{Question selection and information elicitation.}
Psychiatric intake is adaptive, in that the questions an interviewer selects determine what information a patient ultimately reveals. In psychiatric intake, the information revealed, and the questions needed to elicit it, can vary substantially from patient to patient \cite{Oneilll2021}. Recent works frame this process as a sequential decision problem \cite{li2024, li2025, gui2026}, where an LLM must choose the next best question under uncertainty and a limited budget. As more AI intake systems are developed, health systems need efficient quality assurance processes that reflect their clinical practices.

\paragraph{Simulated and standardized patients.}
Standardized patients played by trained actors are a common approach for assessing clinical interviews. Yet, in studies using standardized patients, physicians elicited only around \textbf{55\%} of the information experts judged important, and open-ended questions were substantially stronger predictors of disclosure than closed-ended questions (r = 0.72 vs r = 0.37)~\cite{roter1987}. In addition, standardized patients are resource-intensive, vary in performance, and are often constrained by the local actor pool \cite{flanagan2023}. 

LLM-based simulated patients are a promising alternative, many of which are primarily built as educational training tools for learners. PATIENT-$\Psi$ grounds the simulation in expert-authored cognitive behavioral therapy models \cite{wang2024}, while Roleplay-doh elicits behavioral principles from domain experts \cite{louie2024}, and PatientSim varies persona along four validated axes in emergency-department encounters \cite{kyung2025}. 

InterviewPlayground \cite{ivey2026} is built to assess an interview, making it appropriate for our task. Rather than incorporating case descriptions in the prompt, InterviewPlayground grounds their simulated participants in a bank of retrieved autobiographical memories that correlate strongly with real qualitative interviews on response quality (r = 0.84), interviewer behavior (r = 0.96), and several dimensions of participant experience for qualitative research interviews (r = 0.70) \cite{ivey2026}. We adapt its retrieval mechanism for psychiatric intake to simulate psychiatric patients as part of our approach. 

\paragraph{Adaptive, task-Specific evaluations.}
Recent work has begun moving AI x psychiatric evaluation beyond static, full-information settings. Safe-Psych, for example, reveals clinical evidence sequentially and evaluates whether a model appropriately diagnoses, seeks clarification, or abstains as information accumulates \cite{presacan2026}. This tests whether a model recognizes when more information is needed, but the model does not itself decide what questions to ask or what information to elicit.
% Similarly, head-to-head comparisons between clinicians and conversational AI have also been conducted in general medicine \cite{tu2025}, but primarily assess diagnostic accuracy rather than the quality of the interview process.

Our platform extends this adaptive framing for the intake process itself: whether an interviewer elicits clinically important information and responds appropriately to what the simulated patient reveals.

\section{Participatory Design}

Our platform was developed by a ten-person interdisciplinary team comprising five members with CS and AI expertise and five behavioral-health professionals and trainees. The team met weekly, and developed an analysis plan and a study pre-registration (10.17605/OSF.IO/54XEJ) before data collection. Design decisions were either piloted or voted on.

At each step, there were important design considerations. First, comparisons across interviewing styles required simulated patients that were repeatable enough for controlled evaluation while still supporting adaptive conversation. Second, clinician participation had to be brief enough to be practical while preserving a realistic intake process. Third, information recovery provided a simple quantitative comparison, but could not capture unsupported inference or inadequate responses to safety concerns. These design considerations correspond to our approach for the synthetic patient, simulated intake platform, and intake evaluation components, as described below.

\section{Method}
As shown in Figure~\ref{fig:pipeline}, the platform has three main components: the simulated patient, the simulated intake platform, and the intake evaluation. %% Each component corresponds to one of the three design constraints above. 

\subsection{1. Simulated Patient Construction}
\begin{figure}[h]
\centering
\includegraphics[width=\columnwidth]{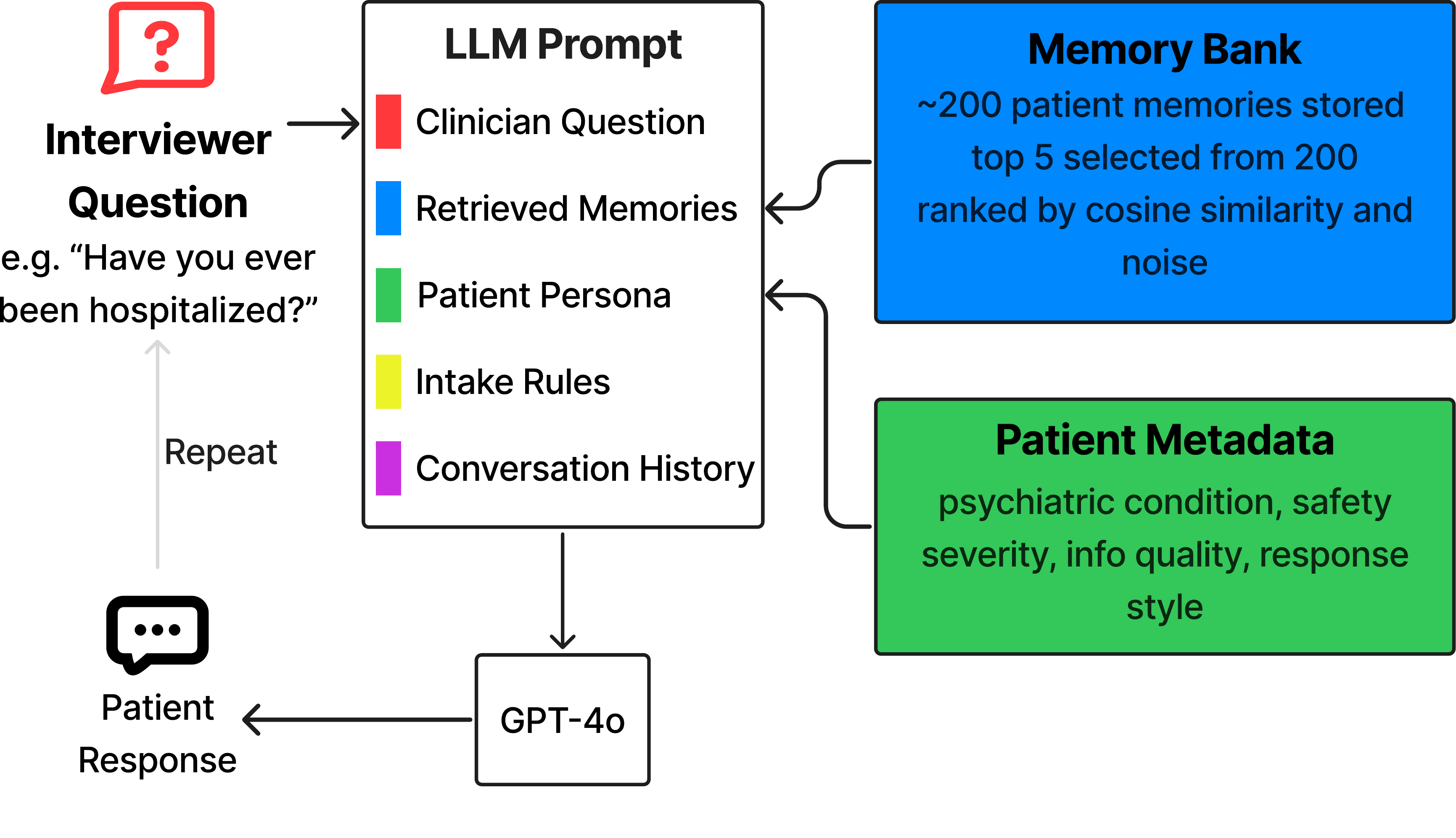} 
\caption{Patient simulator interview loop. Each question triggers memory retrieval from the memory bank, and the memories are injected into the response prompt alongside the question, patient persona, intake rules, and conversation history before an LLM generates a reply.}
\label{fig:turn-sequence}
\end{figure}

Our clinical and behavioral health team developed 12 psychiatric intake vignettes, three of which were used in this study. They were chosen to span distinct diagnostic presentations based on frequency and difficulty and included 1) major depression with suicidal ideation, 2) generalized anxiety, and 3) bipolar II disorder. Each vignette contains demographic information, a short patient description, communication style (verbosity, information quality, disclosure), an embedded safety concern, and ten binary evaluation fields that serve as ground truth for scoring. The fields cover family psychiatric history, employment, living situation, relationship status, children, prior psychiatric treatment, and current use of alcohol, tobacco, cannabis, and other substances. Because every vignette carries the same ten fields, interviews of different synthetic patients are scored against the same targets and remain comparable. Additionally, the safety concern was an intentionally included safety-relevant disclosure that should prompt further clinical assessment. 

\textit{Design Consideration:} Our first synthetic patient consisted of a system prompt that included the patient vignette, the extended history and communication style, behavioral instructions, and the recent conversation history, and instruction to remain in character. In pilot testing, the synthetic patient invented and revised medical history during the interview and occasionally failed to respond as the conversation progressed. These drawbacks motivated the adoption of a memory-augmented retrieval architecture from InterviewPlayground \citep{ivey2026}.

To implement the memory-augmented retrieval architecture, each vignette was converted into a persona description and a trait vector and allocated roughly 200 memory slots, ten of which were target memories, with one per evaluation field. The remaining 190 slots were generated by an LLM and featured concrete sensory details. Memory generation happened once, offline, and the result was committed as a preset file and used in every subsequent interview.  %% Every clinician therefore interviews the same patient, and the fact a clinician could have elicited is the same object the model draws on when answering. 
In addition, we added intake-specific patient behavior, such as patient not volunteering self-harm content without direct and careful questioning \cite{hallfordDisclosureSuicidalIdeation2023} and minimizing substance use disclosure at the first ask while disclosing ordinary life history with concrete lived details.

% The fields are listed here: . %% Because every vignette carries the same ten fields, an interview of any patient can be scored against the same targets, and interviews of different patients remain comparable.

During the simulated intake, each interviewer message triggered memory retrieval before generation (Fig.~\ref{fig:turn-sequence}). The message was embedded and scored by cosine similarity against every memory, with a Gaussian noise term added. The five highest-scoring memories were injected into the response prompt as impressions the synthetic patient was instructed to paraphrase rather than recite.

% The vignette metadata determines how this system behaves. \texttt{response\_style} sets verbosity, \texttt{info\_quality} sets both the patient's memory trait and the noise standard deviation ($0$, $0.075$, and $0.15$ at high, medium, and low), and communication-style descriptions containing guarded language lower the disclosure trait. Three further keys, \texttt{condition}, \texttt{severity}, and \texttt{safety\_severity}, record what the case represents for analysis and do not affect behavior.

 %% This asymmetry keeps the patient realistic while making elicitation a non-trivial task.

\subsection{2. Simulated Intake Platform}

%% Clinician time is an important constraint on a study like this. An interview must be long enough for a clinician to conduct intake normally, but short enough that they actually complete the study. The simulated intake must also be simple enough that clinicians can complete it asynchronously. 

\emph{Design Consideration:} We initially assigned three eight-minute interviews per clinician to capture intake behavior across multiple synthetic patient presentations. However, two separate pilot sessions with medical students and clinicians (n=35) showed that eight minutes was insufficient. Clinicians frequently ran out of time or would stop the study at the second or third interview. 
The final design was a single 20-minute interview per clinician in order to preserve clinical realism while keeping the interview manageable. Pilot sessions also drove two accessibility changes: i) attaching a brief scripted tutorial demonstrating the chat interface at the beginning of the experiment, ii) adding a short delay before simulated patient replies for realism. 

We recruited clinicians with psychiatric intake experience by email through departmental networks. Of 21 who consented, 12 completed the full study, and seven remained after exclusion criteria as follows: non-clinician status, no recall form completed, fewer than five messages sent, or a failed attention check (a simple question about the patient case).

% Two services support this step. A React and Express application backed by a hosted PostgreSQL database handles consent, assignment, session timing, transcript capture, and the recall instrument, and streams patient replies as they are generated (Fig.~\ref{fig:platform}). A separate Python service exposes the simulated patient over an internal HTTP interface.

\begin{figure}[h]
\centering
\includegraphics[width=\columnwidth]{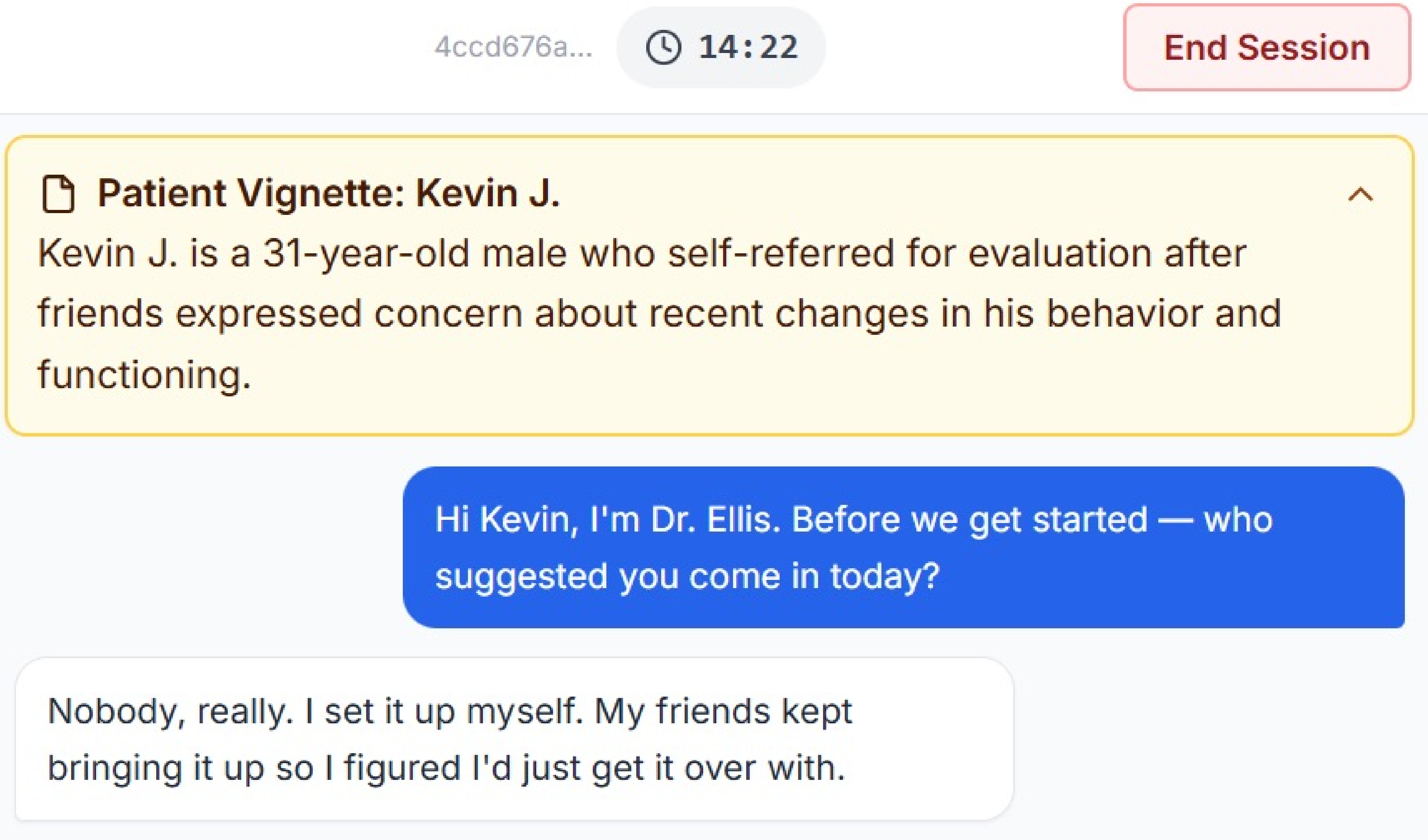}
\caption{Simulated intake platform interface includes a small snippet of the patient information and includes a timer for the interviewer to perform intake.}
\label{fig:platform}
\end{figure}

%% Evaluating an intake requires knowing what was elicited during the conversation and what the interviewer took away. A post-interview form records what the interviewer believes, but cannot separate a topic never raised from one raised and forgotten, or from one answered confidently on no evidence. A transcript records what was said, but not what the interviewer concluded from it. 

\paragraph{Recall Form:} 
After the intake process, clinicians were directed to a separate screen and completed a recall and feedback form administered immediately after the interview with three parts:

\begin{itemize}
    \item Three open-ended questions asking for the patient's chief complaint, any identified safety concerns, and the clinical thread being pursued.
    \item Likert-scale items (\emph{Definitely No, Probably No, Unclear/Not Addressed, Probably Yes, Definitely Yes}) covering specific clinical topics. Parent questions revealed follow-up items on personal/social history and substance use, and this section included a vignette-specific attention check.
    \item Questions on patient realism and tool usability.
\end{itemize}

\subsection{3. Intake Evaluation }

\textit{Design Consideration:} For quality assurance, we did not want the evaluation to reduce intake quality to a single information-recovery score. An interviewer could recover many clinically relevant fields while still making unsupported inferences or responding inadequately to a safety disclosure. We therefore evaluated information recovery alongside the following dimensions:

\textbf{Extracted Information Recall:} Clinician responses are extracted from the recall form and are scored against the vignette binary evaluation fields. A response of \emph{unclear} or \emph{not addressed} is not a directional answer, so it is excluded from precision and counted as incorrect in accuracy.

Precision is correct responses over the responses an interviewer committed to. Accuracy is correct responses over all ten fields. We compute both because precision alone favors an interviewer who commits rarely, while accuracy alone in this set-up does not distinguish between a wrong answer and a Unclear/Not Addressed.

Interviewer transcripts are scored by having GPT-4o complete the same recall form the clinicians completed, working from the transcript alone with no access to the vignette or ground truth. The form is administered in the same two stages, so a negative answer to a gate question suppresses the same four detail items it suppresses for a clinician.

\textbf{Transcript-Based Responsiveness:} We code two features of interviewer question formulation. The first is whether a question opens with an inverted auxiliary verb (\emph{Have you\ldots}, \emph{Do you\ldots}), which signals a question expecting a yes or no answer. The second is whether a question invites elaboration, meaning it cannot be adequately answered with a single fact. These were evaluated using a LLM-judge approach, labeling each message as open, closed, or not a question.

\textbf{Inferred Information Recall:} We define clinical inference as cases where an interviewer gave a directional answer about a topic the transcript shows was never raised.

\textbf{Safety Concern Identification and Characterization:} Each vignette contains a note describing an embedded safety concern (e.g., high self-harm risk). Free-text responses are scored using a language model against that note for whether the concern was identified, whether appropriate detail was elicited, and whether a concern absent from the vignette was asserted. We report sensitivity metrics as every vignette in the active pool carries a concern (i.e. no true-negative case exists). Identification and characterization are scored separately, since detecting a safety signal and establishing what it means are meaningfully distinct.

\section{Experimental Evaluation and Results}

We evaluated two types of interviewers: behavioral health clinicians and a GPT-4o \footnote{GPT-4o is among the models available to Johns Hopkins users through HopGPT, the institution's secure generative-AI environment approved for handling protected health information (PHI).} intake interviewer. Of the seven eligible clinician interviews (four psychiatrists and three psychologists), one vignette had only a single clinician response and was excluded, leaving six clinician sessions across two vignettes. We compared these with 10 LLM-based  (GPT-4o) interviewers on the same vignettes (five each).

Because clinicians used nearly the full interview window but varied substantially in how many messages they sent (14–37 messages; median duration 20.2 minutes), we matched the LLM interviewer to clinician message counts rather than elapsed time. The LLM received a fixed budget equal to the mean clinician message count for each vignette: 26 and 29 messages, respectively.

%For comparison, the LLM interviewer is a plain GPT-4o wrapper intended to stand in for a basic automated intake chatbot. It receives the same vignette text the clinician saw and an instruction adapted from the clinician's and it is not given the target fields, any question bank, or any indication of what will be scored, and a runtime check verifies that no field name reaches the prompt.

% Each clinician session was scored in two ways, giving three conditions: clinician interviews scored by clinician recall, clinician interviews scored by automated extraction, and LLM interviews scored by automated extraction. The first two conditions differ in documentation mechanism; the second and third differ only in interviewer. 

% ---------- Session-level comparison ----------

\subsection{R1. Simulated patient and intake realism}
Across the 16 interviews, the simulated patient was generally consistent, adhering to the vignette in 18 of the 20 binary evaluation fields, and deviating in only two instances. One patient shared occasional wine use against a profile listing no alcohol use, and another described past psychotherapy in one session and denied it in two others against a profile that had no specification about past therapy.

Clinicians rated the synthetic patients (1 \emph{slightly}, 4 \emph{moderately}, 1 \emph{very}) realistic and the platform as (1 \emph{easy}, 5 \emph{very easy}) to use, with no bug reports. Four clinicians separately described guardedness, minimization, or constricted affect matching the scripted communication style, and none reported a factual inconsistency.

\subsection {R2: Clinician burden and performance}

The median time from registration to completion was 25.3 minutes, approximately 20 of which were spent on the interview. The recall form itself took a median of 2.8 minutes. Clinician messages averaged 15.6 words while the LLM averaged 22.2.  Clinicians also left substantially more evaluation fields unclear or not addressed than the LLM (50\% $\pm$ 26\% vs.\ 22\% $\pm$ 9\%). This indicates that clinicians more often declined to make a directional judgment when the relevant information had not been clearly elicited, whereas the LLM was more likely to commit to an answer.

Clinicians produced messages containing no question in 9\% of messages, and the LLM in 2\%. On inspection each of the six LLM messages still contained a question preceded by a greeting or acknowledgment. No LLM message in any run lacked a question. This matters because questions aren't the only way to elicit information. One clinician comment: that the patient could not fight their way out of the numbness, drew a fuller disclosure than the question preceding it.

\subsection{R3. Intake Evaluation Results}

% \paragraph{R3. Information Recall and Evaluation}

\textbf{Extracted Information Recall:} Clinicians' recall form and the transcript-extracted one agreed on 87.1\% of binary evaluation fields. For consistency, we compare the clinician's interview transcript and the LLM's interview transcript in this section.

\begin{table}[h]
\centering
\small
\caption{Mean session-level comparison between extractions from clinician interview transcripts ($n=6$) and LLM transcripts ($n=10$) $\pm$ 95\% CI.}
\label{tab:session_results}
\begin{tabular}{lcc}
\toprule
\textbf{Measure} & \textbf{Clinician} & \textbf{LLM} \\
\midrule
Precision                         
& $\mathbf{100\% \pm 0\%}$ & $95.97\% \pm 7\%$ \\
Accuracy                         
& $50\% \pm 26\%$ & $\mathbf{74\% \pm 11\%}$ \\
Polar-question openings          
& $34\% \pm 27\%$ & $\mathbf{77\% \pm 10\%}$ \\
Questions inviting elaboration   
& $27\% \pm 18\%$ & $\mathbf{41\% \pm 9\%}$ \\
\bottomrule
\end{tabular}
\end{table}

% Clinicians and LLMs diverged most on the fields where a fact had to be surfaced rather than assumed absent (positive ground-truths) . Clinicians did not answer incorrectly: all 30 of their directional responses were correct. Of the 11 fields they missed, seven were answered \emph{unclear or not addressed} and four were never shown, because they were gated by a negative answer to the parent question on the Likert scale. Clinician precision is therefore 100\%. 

% Asking two things at once interfered with extraction. Both incorrect extractions of prior psychiatric treatment followed this pattern:

% \begin{quote}\small
% \textbf{LLM:} Have you ever spoken to a doctor or therapist about your
% anxiety before, or is this your first time seeking professional support?

% \textbf{Patient:} This is my first time. \ldots
% \end{quote}

% \noindent The patient answered unambiguously, but the extraction step
% recorded prior treatment as present. Both runs asking the same question
% as a single question were coded correctly.

\textbf{Transcript-Based Responsiveness:} As shown in Table \ref{tab:session_results}, two measures of openness point in opposite directions. The LLM opened questions with inverted auxiliaries more than twice as often as clinicians did, yet a higher proportion of its questions were coded as inviting elaboration. The LLM tended to ask two things at once, opening with a yes-or-no question and then attaching an alternative that gives the simulated patient somewhere to go: \emph{Have you ever spoken to a doctor about your anxiety, or is this your first time seeking support?} The question begins with a closed yes/no form, so it is counted as an inverted-auxiliary opening, but the added alternative broadens the response space and is classified as inviting elaboration.

\begin{table}[h]
\centering
\small
\caption{Performance across clinician sessions and LLM runs.}
\label{tab:pooled_results}
\begin{tabular}{lcc}
\toprule
\textbf{Outcome} & \textbf{Clinician} & \textbf{LLM} \\
\midrule
Positive vignette fields elicited 
    & $38.9\%$ (7/18)
    & $\mathbf{88.0\%}$ (22/25) \\

% \quad Four-positive sessions
%     & $31.3\%$ (5/16)
%     & $\mathbf{85.0\%}$ (17/20) \\

% \quad One-positive sessions
%     & $\mathbf{100\%}$ (2/2)
%     & $\mathbf{100\%}$ (5/5) \\

Directional response accuracy
    & $\mathbf{100\%}$ (30/30)
    & $94.9\%$ (74/78) \\

Incidence of inferred answers
    & $\mathbf{27.8\%}$ (5/18)
    & $56.8\%$ (21/37) \\

Safety feature identified
    & $50.0\%$ (3/6)
    & $\mathbf{90.0\%}$ (9/10) \\

\quad With qualifying detail
    & $\mathbf{66.7\%}$ (2/3)
    & $33.3\%$ (3/9) \\
\bottomrule
\end{tabular}
\end{table}

\paragraph{Inferred Information Recall:}

Both clinicians and LLMs committed to answers on topics the transcript shows were never raised, the LLM ~2.5$\times$ as often per session. Every inference in the study was negative about the characteristic and nearly all were correct, clinicians with 100\%, the LLM 95\% of the time. These inferences were concentrated in fields where a negative is the common answer. For example, family psychiatric history, which is relatively common, was never inferred to be negative. 

Deliberation was rare: across all of the Likert-based responses, we recorded a single revision, in which a clinician withdrew a directional answer about the patient's living situation and replaced it with \emph{unclear or not addressed} eleven seconds later.\\

% ---------- Pooled outcomes ----------
\textbf{Safety Concern Identification and Characterization:} The LLM identified a safety concern in 9/10 runs, against 3/6 clinician sessions, but characterized what it found far less often: 3/9 identifications carried qualifying detail, against 2/3 for clinicians. For the LLMs runs, one did not ask a safety question at all, six received a disclosure and changed topic without acknowledging it,  one acknowledged the disclosure and then changed topic, one followed up on prior episodes rather than on risk, and one probed means for self-harm.

\begin{quote}\small
\textbf{LLM:} Have you been having any thoughts about wanting to harm yourself or feeling like life isn't worth living?

\textbf{Patient:} I don't know\ldots not really. I just get tired of everything sometimes, that's all. \ldots But I'm not going to do anything.

\textbf{LLM:} Have you noticed any changes in your ability to concentrate or make decsions\ldots
\end{quote}

\noindent

A clinician on the same vignette treated a comparable deflection as something to work through, asking across six consecutive turns about anhedonia, outlook, passive ideation, hopelessness, and finally means, and re-asking directly when the patient said it was hard to explain.

\section{Discussion}

The purpose of this platform is to support evaluation of conversational psychiatric intake tools before patient-facing deployment. Our results show that these evaluations should not just output a single score for quality assurance. The LLM recovered more of the ground truth than the clinicians, but also performed worse on quality assurance metrics important to our application. Similarly, while the LLM detected safety concerns more often, it characterized what it found less than half as often. In most runs, the simulated patient disclosed passive suicidal ideation and the LLM changed the subject. Given limited clinician bandwidth, surfacing these higher-level failure patterns may be a more effective use of expert time than repeatedly stress-testing these systems individually. 

\paragraph{Patient comparability}
Our design of the synthetic patients did make cross-interviewer comparisons possible and also highlighted the settings in which it failed. A vignette should specify its negatives as deliberately as its positives, and the clinicians in our study probed exactly the gaps the LLM never reached. In the future, we can also include prior transcripts as the simulation is shown to more clinicians so that it is consistent with prior transcripts, but this would require more compute. Finally, humans are also inconsistent between different interviewers and their histories, and this is a dimension that needs the most consideration before deployment. 

\paragraph{Clinician participation}
Out of the 21 participants who consented, 12 completed the experiment. No one abandoned an interview in progress, but many stopped before the recall form step. This type of attrition is expected in studies like this given the challenging time pressures of a clinical role. Designing a smoother transition and being clearer about the recall form beforehand, like in the demo, as well as additional incentives, could improve clinician completion.

\paragraph{Deployment-relevant evaluation}
Our evaluation approach distinguished information recovery from clinically consequential failure modes. Although the LLM recovered more positive ground-truth information than clinicians (88\% vs. 39\%), the broader evaluation exposed weaknesses in unsupported inference and safety follow-up. This separation is the key design goal for deployment, where health systems need interpretable dimensions of performance rather than a single aggregate score.

% The evaluation exercise generated several discussions about developing thresholds for acceptability for a deployed system. Many thresholds and considerations of metrics to include depend on the culture and context of deployment. For example, while the LLM detected more safety concerns than clinicians, they characterized at a lower rate. Recall precision ranks clinicians above the LLM, but clinician precision is skewed by non-answers, so it's not as valuable of a metric. Accuracy is inflated by interviewer inferences, which is highly detrimental if an AI system is used on a patient for whom these inferences would be inappropriate (e.g., an individual who may not be well represented in the model's training set). Clinical-trial readiness depends on these discussions and using tools like this as a discussion point that needs to be continuously updated, rather than a simple score. 

\section{Pathway to Deployment}

Our objective is to develop an approach that will serve as a consistent tool for quality assurance in AI-assisted psychiatric intake. Before deployment, a hospital would evaluate different candidates against a set of vignettes and criteria relevant to their setting, and compare the outputs from their clinicians and styles to more thoroughly and systematically assess the potential AI intake tools. The same evaluation would then be rerun after major model, prompt, or workflow changes and while it is deployed as standards change. 

To reach this point, there are a few technical steps.  First, the simulated patients themselves require stronger validation against real psychiatric intake. The current study demonstrates consistency across repeated simulated encounters, but does not establish that the simulator reproduces real patient disclosure patterns or the full complexity of clinical interaction (e.g. reacting to inconsistencies or anger). A next step is to validate simulated encounters against real cases, stratified by common presenting concerns and patient groups, and with enough cases in each group to support meaningful comparison. The approach should also be tested across multiple model families. 

Second, the quality assurance approach needs to be validated alongside an actual AI intake system in a clinically supervised setting. We are working on this in tandem, and it requires institutional review, privacy protections, and clinical oversight for patient AI use, particularly because psychiatric intake can involve self-harm, substance use, trauma, and other critical patient-safety disclosures. Autonomous deployment would ultimately require validation in a large-scale clinical trial, and this research establishes a mechanism and results to support such a trial as we pursue this goal. 

\section{Conclusion}

We present a clinician-grounded evaluation platform for AI-assisted psychiatric intake designed to support meaningful comparison across interviewing styles, low-burden clinician participation, and deployment-relevant evaluation. To address these challenges, we created memory-augmented simulated patients with fixed ground truths, a simulated intake platform, and evaluation modalities that include both a recall form and transcript analysis. In a pilot with clinicians from the Hopkins Medicine Department of Psychiatry, the platform enabled quality assurance discussions that information recovery evaluations alone would have missed: an LLM interviewer that reached more of the ground truth than clinicians while rarely following up on the safety concerns it detected. This clinician-grounded framework provides health systems with a practical foundation for supporting ongoing quality assurance for AI-assisted psychiatric intake.

% Beyond psychiatry, the same approach can be useful for wherever a conversational system must be assessed before it can be safely deployed for human users.

% %% ============================================================
% \section*{Ethical Statement}

% %% ============================================================
% \section*{Acknowledgments}

%% ============================================================
%% References do not count against the 6-page limit.
\bibliography{references}

@misc{gui2026,
      title={Optimal Question Selection from a Large Question Bank for Clinical Field Recovery in Conversational Psychiatric Intake}, 
      author={Gui, Guan and Zandi, Peter and Taylor, Jacob and Joshi, Ananya},
      year={2026},
      eprint={2604.22067},
      archivePrefix={arXiv},
      primaryClass={cs.CL},
      url={https://arxiv.org/abs/2604.22067}, 
}

@unpublished{ivey2026,
  author={Ivey, Jonathan and Liang, Aimee and Wang, Y.S Arthur and Mandell, Madeline and Xiao, Ziang and Field, Anjalie},
  title={InterviewPlayground: A Validated Simulation Environment for Evaluating AI Interviewers},
  note={Manuscript in preparation},
  year={2026}
}

@article{flanagan2023,
  title = {Standardized Patients in Medical Education: A Review of the Literature},
  author = {Flanagan, Octavia L and Cummings, Kristina M},
  year = 2023,
  journal = {Cureus},
  volume = {15},
  number = {7},
  pages = {e42027},
  issn = {2168-8184},
  doi = {10.7759/cureus.42027},
  pmcid = {PMC10431693},
  pmid = {37593270}
}

@article{fouda2026,
  title = {{{PsychiatryBench}}: A Multi-Task Benchmark for {{LLMs}} in Psychiatry},
  shorttitle = {{{PsychiatryBench}}},
  author = {Fouda, Aya E. and Hassan, Abdelrahman A. and Hanafy, Radwa J. and Fouda, Mohammed E.},
  year = 2026,
  journal = {NPJ Digital Medicine},
  volume = {9},
  pages = {320},
  issn = {2398-6352},
  doi = {10.1038/s41746-026-02582-w},
  pmcid = {PMC13087022},
  pmid = {41981155}
}

@article{gallifant2025,
  title = {The TRIPOD-LLM reporting guideline for studies using large language models},
  author = {Gallifant, Jack and Afshar, Majid and Ameen, Saleem and Aphinyanaphongs, Yindalon and Chen, Shan and Cacciamani, Giovanni and {Demner-Fushman}, Dina and Dligach, Dmitriy and Daneshjou, Roxana and Fernandes, Chrystinne and Hansen, Lasse Hyldig and Landman, Adam and Lehmann, Lisa and McCoy, Liam G. and Miller, Timothy and Moreno, Amy and Munch, Nikolaj and Restrepo, David and Savova, Guergana and Umeton, Renato and Gichoya, Judy Wawira and Collins, Gary S. and Moons, Karel G. M. and Celi, Leo A. and Bitterman, Danielle S.},
  year = 2025,
  journal = {Nature Medicine},
  volume = {31},
  number = {1},
  pages = {60--69},
  publisher = {Nature Publishing Group},
  issn = {1546-170X},
  doi = {10.1038/s41591-024-03425-5},
  copyright = {2025 The Author(s), under exclusive licence to Springer Nature America, Inc.}
}

@article{hua2025,
  title = {Charting the Evolution of Artificial Intelligence Mental Health Chatbots from Rule-based Systems to Large Language Models: A Systematic Review},
  author = {Hua, Yining and Siddals, Steve and Ma, Zilin and Galatzer-Levy, Isaac and Xia, Winna and Hau, Christine and Na, Hongbin and Flathers, Matthew and Linardon, Jake and Ayubcha, Cyrus and Torous, John},
  year = 2025,
  journal = {World Psychiatry},
  volume = {24},
  number = {3},
  pages = {383--394},
  issn = {1723-8617},
  doi = {10.1002/wps.21352},
  pmcid = {PMC12434366},
  pmid = {40948070}
}

@misc{kyung2025,
  title = {PatientSim: A Persona-Driven Simulator for Realistic Doctor-Patient Interactions},
  author = {Kyung, Daeun and Chung, Hyunseung and Bae, Seongsu and Kim, Jiho and Sohn, Jae Ho and Kim, Taerim and Kim, Soo Kyung and Choi, Edward},
  year = 2025,
  number = {arXiv:2505.17818},
  eprint = {2505.17818},
  primaryclass = {cs.AI},
  publisher = {arXiv},
  doi = {10.48550/arXiv.2505.17818},
  archiveprefix = {arXiv},
}

@misc{louie2024,
  title={Roleplay-doh: Enabling Domain-Experts to Create LLM-simulated Patients via Eliciting and Adhering to Principles}, 
  author={Ryan Louie and Ananjan Nandi and William Fang and Cheng Chang and Emma Brunskill and Diyi Yang},
  year={2024},
  eprint={2407.00870},
  archivePrefix={arXiv},
  primaryClass={cs.CL},
  url={https://arxiv.org/abs/2407.00870}, 
}

@inproceedings{moore2025,
  title = {Expressing Stigma and Inappropriate Responses Prevents {{LLMs}} from Safely Replacing Mental Health Providers},
  booktitle = {Proceedings of the 2025 {{ACM Conference}} on {{Fairness}}, {{Accountability}}, and {{Transparency}}},
  author = {Moore, Jared and Grabb, Declan and Agnew, William and Klyman, Kevin and Chancellor, Stevie and Ong, Desmond C. and Haber, Nick},
  year = 2025,
  eprint = {2504.18412},
  primaryclass = {cs.CL},
  pages = {599--627},
  doi = {10.1145/3715275.3732039},
  archiveprefix = {arXiv}
}

@article{pichowicz2025,
  title = {Performance of Mental Health Chatbot Agents in Detecting and Managing Suicidal Ideation},
  author = {Pichowicz, W. and Kotas, M. and Piotrowski, P.},
  year = 2025,
  journal = {Scientific Reports},
  volume = {15},
  number = {1},
  pages = {31652},
  publisher = {Nature Publishing Group},
  issn = {2045-2322},
  doi = {10.1038/s41598-025-17242-4},
  copyright = {2025 The Author(s)}
}

@misc{presacan2026,
  title = {Ask {{Before You Diagnose}}: {{Safe-Psych}}, a {{Sequential Evaluation Benchmark}} for {{LLMs}} in {{Psychiatry}}},
  shorttitle = {Ask {{Before You Diagnose}}},
  author = {Presacan, Oriana and Grama, Andreea and Irimin{\u a}, Larisa and Nik, Alireza and Ojha, Jaya and Thambawita, Vajira and B{\u a}cil{\u a}, Ciprian I. and Ionescu, Bogdan and Riegler, Michael A.},
  year = 2026,
  number = {arXiv:2607.13036},
  eprint = {2607.13036},
  primaryclass = {cs.CL},
  publisher = {arXiv},
  doi = {10.48550/arXiv.2607.13036},
  archiveprefix = {arXiv}
}

@article{roter1987,
  title = {Physicians' Interviewing Styles and Medical Information Obtained from Patients},
  author = {Roter, Debra L. and Hall, Judith A.},
  year = 1987,
  journal = {Journal of General Internal Medicine},
  volume = {2},
  number = {5},
  pages = {325--329},
  issn = {1525-1497},
  doi = {10.1007/BF02596168}
}

@article{vasey2022,
  title = {Reporting Guideline for the Early-Stage Clinical Evaluation of Decision Support Systems Driven by Artificial Intelligence: {{DECIDE-AI}}},
  shorttitle = {Reporting Guideline for the Early-Stage Clinical Evaluation of Decision Support Systems Driven by Artificial Intelligence},
  author = {Vasey, Baptiste and Nagendran, Myura and Campbell, Bruce and Clifton, David A. and Collins, Gary S. and Denaxas, Spiros and Denniston, Alastair K. and Faes, Livia and Geerts, Bart and Ibrahim, Mudathir and Liu, Xiaoxuan and Mateen, Bilal A. and Mathur, Piyush and McCradden, Melissa D. and Morgan, Lauren and Ordish, Johan and Rogers, Campbell and Saria, Suchi and Ting, Daniel S. W. and Watkinson, Peter and Weber, Wim and Wheatstone, Peter and McCulloch, Peter},
  year = 2022,
  journal = {Nature Medicine},
  volume = {28},
  number = {5},
  pages = {924--933},
  publisher = {Nature Publishing Group},
  issn = {1546-170X},
  doi = {10.1038/s41591-022-01772-9},
  copyright = {2022 The Author(s), under exclusive licence to Springer Nature America, Inc.}
}

@inproceedings{wang2024,
  title = {{{PATIENT-$\psi$}}: {{Using Large Language Models}} to {{Simulate Patients}} for {{Training Mental Health Professionals}}},
  shorttitle = {{{PATIENT-$\psi$}}},
  booktitle = {Proceedings of the 2024 {{Conference}} on {{Empirical Methods}} in {{Natural Language Processing}}},
  author = {Wang, Ruiyi and Milani, Stephanie and Chiu, Jamie C. and Zhi, Jiayin and Eack, Shaun M. and Labrum, Travis and Murphy, Samuel M and Jones, Nev and Hardy, Kate V and Shen, Hong and Fang, Fei and Chen, Zhiyu},
  editor = {{Al-Onaizan}, Yaser and Bansal, Mohit and Chen, Yun-Nung},
  year = 2024,
  pages = {12772--12797},
  publisher = {Association for Computational Linguistics},
  address = {Miami, Florida, USA},
  doi = {10.18653/v1/2024.emnlp-main.711}
}

@article{RAY2023,
    title = {Benchmarking, ethical alignment, and evaluation framework for conversational AI: Advancing responsible development of ChatGPT},
    journal = {BenchCouncil Transactions on Benchmarks, Standards and Evaluations},
    volume = {3},
    number = {3},
    pages = {100136},
    year = {2023},
    issn = {2772-4859},
    doi = {https://doi.org/10.1016/j.tbench.2023.100136},
    author = {Partha Pratim Ray}
}

@article{Oneilll2021,
  title = {Uncovering the {{Intricacies}} of the {{Clinical Intake Assessment}}: {{How Clinicians Prioritize Information}} in {{Complex Contexts}}},
  author = {O'Neill, Margaret M. and Nakash, Ora},
  year = 2021,
  journal = {Journal of the Society for Social Work and Research},
  volume = {12},
  number = {4},
  pages = {803--829},
  issn = {2334-2315, 1948-822X},
  doi = {10.1086/715439}
}

@misc{li2025,
  title = {{{ALFA}}: {{Aligning LLMs}} to {{Ask Good Questions A Case Study}} in {{Clinical Reasoning}}},
  author = {Li, Shuyue Stella and Mun, Jimin and Brahman, Faeze and Hosseini, Pedram and Thomas, Bryceton G. and Sin, Jessica M. and Ren, Bing and Ilgen, Jonathan S. and Tsvetkov, Yulia and Sap, Maarten},
  year = 2025,
  number = {arXiv:2502.14860},
  eprint = {2502.14860},
  primaryclass = {cs.CL},
  publisher = {arXiv},
  doi = {10.48550/arXiv.2502.14860},
  archiveprefix = {arXiv},
}

@inproceedings{li2024,
  title = {{{MediQ}}: {{Question-Asking LLMs}} and a {{Benchmark}} for {{Reliable Interactive Clinical Reasoning}}},
  booktitle = {Advances in {{Neural Information Processing Systems}}},
  author = {Li, Shuyue Stella and Balachandran, Vidhisha and Feng, Shangbin and Ilgen, Jonathan S. and Pierson, Emma and Koh, Pang Wei and Tsvetkov, Yulia},
  year = 2024,
  volume = {37},
  pages = {28858--28888},
  publisher = {Curran Associates, Inc.},
  doi = {10.52202/079017-0908}
}

@misc{liu2025,
  title={PsychBench: A comprehensive and professional benchmark for evaluating the performance of LLM-assisted psychiatric clinical practice}, 
  author={Shuyu Liu and Ruoxi Wang and Ling Zhang and Xuequan Zhu and Rui Yang and Xinzhu Zhou and Fei Wu and Zhi Yang and Cheng Jin and Gang Wang},
  year={2025},
  eprint={2503.01903},
  archivePrefix={arXiv},
  primaryClass={cs.CL},
  url={https://arxiv.org/abs/2503.01903}, 
}

@article{silverman2015,
    author = {Joel J. Silverman and Marc Galanter and Maga Jackson-Triche and Douglas G. Jacobs and James W. Lomax and Michelle B. Riba and Lowell D. Tong and Katherine E. Watkins and Laura J. Fochtmann and Richard S. Rhoads and Joel Yager},
    title = {The American Psychiatric Association Practice Guidelines for the Psychiatric Evaluation of Adults},
    journal = {American Journal of Psychiatry},
    volume = {172},
    number = {8},
    pages = {798-802},
    year = {2015},
    doi = {10.1176/appi.ajp.2015.1720501},
    URL = {https://psychiatryonline.org/doi/abs/10.1176/appi.ajp.2015.1720501},
    eprint = {https://psychiatryonline.org/doi/pdf/10.1176/appi.ajp.2015.1720501}
}

@article{hirschfeld2003,
  title = {Perceptions and Impact of Bipolar Disorder: How Far Have We Really Come? Results of the National Depressive and Manic-Depressive Association 2000 Survey of Individuals With Bipolar Disorder},
  author = {Robert M. A. Hirschfeld and Lana A. Vornik and Lydia Lewis},
  journal = {The Journal of Clinical Psychiatry},
  volume = {64},
  year = {2003},
  number = {2},
  pages = {14089},
}

@article{hallfordDisclosureSuicidalIdeation2023,
  title = {Disclosure of Suicidal Ideation and Behaviours: {{A}} Systematic Review and Meta-Analysis of Prevalence},
  author = {Hallford, D. J. and Rusanov, D. and Winestone, B. and Kaplan, R. and {Fuller-Tyszkiewicz}, M. and Melvin, G.},
  year = 2023,
  journal = {Clinical Psychology Review},
  volume = {101},
  pages = {102272},
  issn = {0272-7358},
  doi = {10.1016/j.cpr.2023.102272},
}

@techreport{hrsa2026hpsa,
  author = {{Health Resources and Services Administration}},
  title = {Designated Health Professional Shortage Areas Statistics: Fourth Quarter of Fiscal Year 2026 Designated HPSA Quarterly Summary},
  institution = {Bureau of Health Workforce, U.S. Department of Health and Human Services},
  year = 2026,
  url = {https://data.hrsa.gov/topics/health-workforce/shortage-areas},
}

@misc{apa2025advisory,
  author = {{American Psychological Association}},
  title = {APA Health Advisory on the Use of Generative AI Chatbots and Wellness Applications for Mental Health},
  howpublished = {American Psychological Association},
  year = {2025},
  url = {https://www.apa.org/topics/artificial-intelligence-machine-learning/health-advisory-chatbots-wellness-apps},
}

%% ============================================================
%% TECHNICAL APPENDIX
%% No page limit. 
%% Nothing load-bearing goes here. The 6-page body must stand alone.
%% Contents:
%%   - Full metric formulas and index definitions
%%   - Recall instrument: complete item list and gating logic
%%   - Empathy coding rubric and inter-rater procedure
%%   - Vignette schema and the ten evaluation fields
%%   - Memory-generation prompts and retrieval parameters
%%     (top-k, sigma by info_quality level), with justification
%%   - LLM condition: model versions, temperature, time-matching
%%   - Non-significant and exploratory analyses, clearly labeled
%%   - Deviations from the preregistration, if any
%%   - Optional: prompt-only vs. memory-augmented patient comparison
%%   - AAAI reproducibility checklist, if required by the track
%% ============================================================

\end{document}